# A Comparative Study of Machine Learning Models for Hourly Forecasting of Air Temperature and Relative Humidity

Jiaqi Dong

*Abstract*— Accurate short-term forecasting of air temperature and relative humidity is critical for urban management, especially in topographically complex cities such as Chongqing, China. This study compares seven machine learning models: eXtreme Gradient Boosting (XGBoost), Random Forest, Support Vector Regression (SVR), Multi-Layer Perceptron (MLP), Decision Tree, Long Short-Term Memory (LSTM) networks, and Convolutional Neural Network (CNN)-LSTM (CNN-LSTM), for hourly prediction using real-world open data. Based on a unified framework of data preprocessing, lag-feature construction, rolling statistical features, and time-series validation, the models are systematically evaluated in terms of predictive accuracy and robustness. The results show that XGBoost achieves the best overall performance, with a test mean absolute error (MAE) of 0.302 °C for air temperature and 1.271% for relative humidity, together with an average $R^2$ of 0.989 across the two forecasting tasks. These findings demonstrate the strong effectiveness of tree-based ensemble learning for structured meteorological time-series forecasting and provide practical guidance for intelligent meteorological forecasting in mountainous cities.

## I. INTRODUCTION

Accurate short-term forecasting is of great significance for safeguarding the safe operation of cities, improving the efficiency of public services, and responding to extreme weather events. Among various meteorological elements, air temperature and relative humidity are key indicators reflecting the thermal and water vapor status of the near-surface atmosphere, and their hourly variations directly affect energy load scheduling, traffic management, agricultural activities, human comfort, and public health. As a typical mountain-valley city in southwest China, Chongqing has complex local circulations due to its special topography and subtropical monsoon climate. This poses great challenges for traditional numerical weather prediction models in capturing the high spatiotemporal resolution evolution characteristics of temperature and humidity.

In recent years, with the continuous improvement of meteorological observation networks and the rapid development of big data technology, machine learning methods driven by historical observation data have provided a new solution for short-term meteorological forecasting [1], [2]. Compared with traditional models relying on physical equations, data-driven models can automatically learn nonlinear dynamic relationships from massive time-series data and achieve efficient and low-cost prediction. A range of machine learning and deep learning models has shown strong potential in this field. For example, eXtreme Gradient Boosting (XGBoost) [3] has demonstrated high accuracy and robustness for structured tabular data, Long Short-Term Memory (LSTM) networks [4] are well suited for capturing long-term temporal dependencies, and Convolutional Neural Network (CNN)-LSTM (CNN-LSTM) architectures [5] aim to combine local feature extraction with sequential modeling for complex environmental forecasting tasks.

Using hourly observations of air temperature and relative humidity in Chongqing from September 9, 2024, to January 13, 2026, this study conducts a systematic comparison of multiple forecasting models, including classic shallow models (e.g., Support Vector Regression (SVR), Decision Tree (DT)), ensemble methods (e.g., Random Forest, XGBoost), and advanced deep learning architectures (LSTM, CNN-LSTM). By designing rational feature engineering strategies such as lag features, sliding statistical windows, and periodic encoding, and adopting a rigorous time-series cross-validation framework to evaluate model performance, this study aims to screen out the high-precision short-term temperature and humidity forecasting scheme most suitable for Chongqing under its complex geographical and climatic background. The results provide a practical reference for intelligent meteorological prediction in mountain cities with complex geographical and climatic backgrounds.

## II. DATASET

The data used in this study were obtained from publicly accessible websites https://www.visualcrossing.com. Data consists of hourly meteorological observation records of Chongqing from September 9, 2024, to January 13, 2026, covering a complete seasonal cycle with favorable representativeness and seasonal diversity. The dataset includes key meteorological variables: air temperature (0~40 °C) with a bimodal probability density distribution (peaks at 15-20 °C and 25-30 °C) and relative humidity with a right-skewed distribution (peak at 80%-100%), consistent with Chongqing's "mountain city" and "fog city" characteristics. In addition, the dataset contains auxiliary variables such as precipitation (precip), wind speed (windspeed), sea level pressure, cloud cover, and solar radiation, together forming a multi-dimensional and high-resolution meteorological observation system. Exploratory time series analysis reveals that air temperature presents distinct annual periodicity and diurnal variation patterns, whereas humidity shows strong short-term volatility and is closely correlated with precipitation events. As shown in **Fig. 1** and **Fig. 2**, the two target variables exhibit notably different distributional characteristics. The temperature histogram shows a broad and non-uniform distribution, with a pronounced concentration in the lower-temperature range and a gradual spread across higher values. In contrast, the relative humidity histogram is strongly right-skewed, indicating that high-humidity conditions occur most frequently.

Jiaqi Dong is with the Macao Polytechnic University, Macau, China, (corresponding author; e-mail: jiaqidong377@gmail.com).

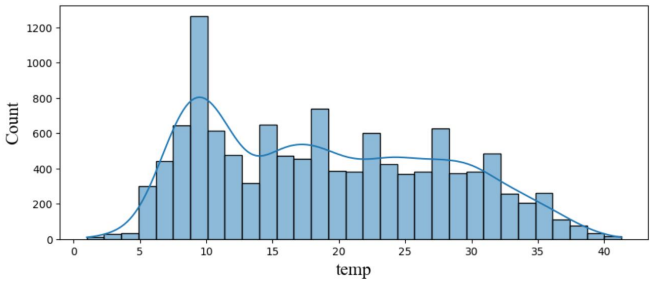

Figure 1.  Frequency distribution plot of hourly temperature observations.

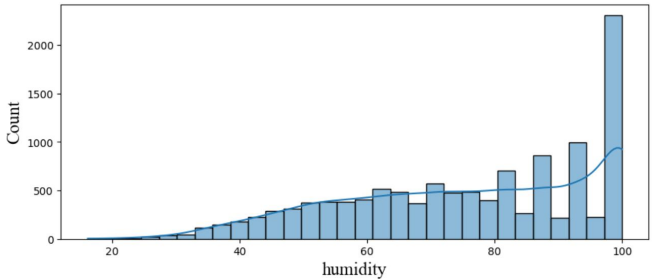

Figure 2.  Frequency distribution plot of hourly relative humidity.

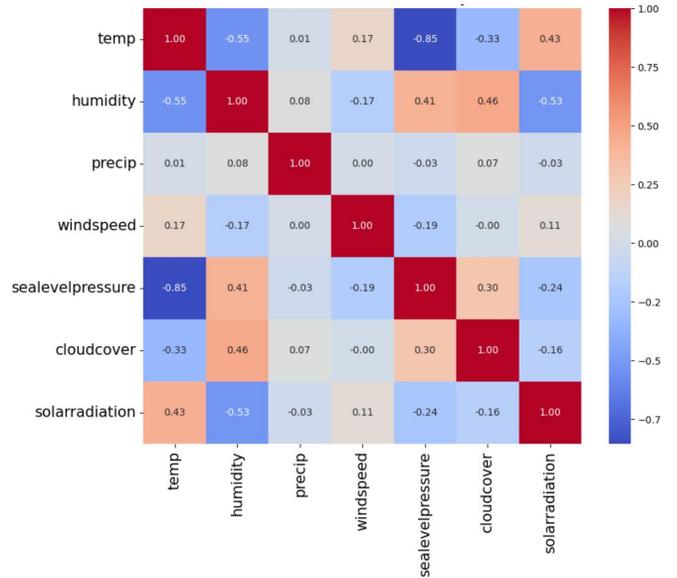

Figure 3.  Weather feature correlation heatmap.

## III. METHODOLOGY

Based on the hourly meteorological observation dataset of Chongqing from September 9, 2024, to January 13, 2026, this study conducts a comparative experiment across multiple forecasting models. To ensure that all models are evaluated on a unified and reliable basis, a systematic workflow consisting of data preprocessing, feature engineering, model construction, and performance evaluation was implemented.

### A. Data Preprocessing

Firstly, the "datetime" field is parsed into timestamps, and the records are strictly sorted in chronological order to ensure the temporal consistency of the time series. For the possible few missing values in the data, a combination of forward filling and backward filling was adopted to maintain continuity in the hourly sequence. Subsequently, a preliminary statistical check is conducted on all variables, and obvious outliers caused by sensor malfunctions or transmission anomalies (such as negative humidity and abnormal temperatures) are removed. Eventually, a high-quality observation sequence with a complete structure and physical rationality is formed, providing reliable input for subsequent modeling.

To further examine the relationships among meteorological variables, Pearson correlation coefficients were computed for all variables and visualized using a correlation heatmap, as shown in **Fig. 3**. The results reveal several clear dependencies among the variables, particularly the strong negative correlation between temperature and sea-level pressure, the negative correlation between temperature and humidity, and the positive correlations of humidity with cloud cover and sea-level pressure.

### B. Feature Engineering

Based on a unified data foundation, this paper constructs a composite feature system for time series prediction around the prediction target. Specifically, for each target variable, two types of time series derived features are systematically generated: lag features, which introduce historical observations from 2, 3, 6, 12, and 24 hours ago to capture the short-term autoregressive dependencies of the variable itself; and rolling statistics features, which calculate the moving average and standard deviation with window lengths of 3, 6, 12, and 24 hours to represent local trend changes and volatility intensity.

In addition, the original auxiliary meteorological variables, including precipitation, wind speed, sea-level pressure, cloud cover, and solar radiation, were retained as external covariates. After combining all derived features with the original variables, a high-dimensional input representation was obtained. Because lag and rolling operations inevitably produce missing values at the beginning of the series, the corresponding samples were removed to ensure a complete modeling dataset. This feature-engineering scheme captures temporal dependence, local dynamics, and multivariable coupling, and provides a common information basis for all forecasting models considered in this study.

### C. Model Construction

To compare different learning paradigms for hourly forecasting, this study constructed seven models: SVR, Multilayer Perceptron (MLP), Random Forest (RF), DT, LSTM, CNN-LSTM, and XGBoost. Since the task involves predicting two target variables, air temperature and relative humidity, simultaneously, different multi-output strategies were adopted according to model characteristics. *MultiOutputRegressor* was used for SVR and XGBoost, whereas DT, RF, and MLP were implemented directly for dual-target regression. For LSTM and CNN-LSTM, the output layer contained two neurons. The input features for each model included raw meteorological variables as well as their derived lag features (2-24 hours) and rolling statistics (mean and standard deviation with a window of 3-24 hours), which fully characterize the autocorrelation and local dynamic characteristics of the time series. For these machine learning models, hyperparameters were optimized using grid search combined with time-series cross-validation

TimeSeriesSplit ($k$=5) to avoid temporal information leakage, and the average RMSE of the two targets was used for model selection.

### a) Support Vector Regression (SVR)

The SVR model was adopted as a kernel-based nonlinear regression model. Since the standard SVR implementation does not natively support multi-output regression, it was wrapped with *MultiOutputRegressor*. The model adopted the Radial Basis Function (RBF) as the kernel function, and systematically optimized the key hyperparameters, including the regularization parameter $C \in \{0.1, 1, 10, 100\}$, the kernel coefficient $\gamma \in \{0.001, 0.01, 0.1, 1\}$, and the insensitive loss tube width $\varepsilon \in \{0.01, 0.1, 0.2\}$.

### b) Multi-Layer Perceptron (MLP)

MLP was implemented using *MLPRegressor* to model nonlinear relationships between the multiple input variables and the two target variables. The model was trained using the Adam optimizer and employed an early stopping mechanism to prevent overfitting and improve generalization. For model optimization, various hyperparameter configurations were explored: The hidden-layer structure was varied among [50], [100], [50, 50], and [100, 100]; the L2 regularization coefficient $\alpha$ was selected from $\{0.0005, 0.001, 0.002\}$; and the learning rate was selected from $\{0.001, 0.005, 0.01\}$, with the maximum iteration number set to 1500. Hyperparameter tuning was performed using a grid search strategy.

### c) Random Forest (RF)

RF was implemented using *RandomForestRegressor* as an ensemble model based on bagging. By aggregating multiple decision trees, it improves robustness and reduces overfitting compared with a single-tree model. In terms of model optimization, a grid search combined with time series cross-validation was applied to perform The main tuned hyperparameters included the number of trees (*n_estimators* $\in \{10, 50, 100\}$), the maximum proportion of features considered for each split (*max_features* $\in \{0.3, 0.5, 0.7\}$), the minimum number of samples required for a leaf node (*min_samples_leaf* $\in \{1, 2, 4\}$), and whether bootstrap sampling is enabled ({True, False}).

### d) Decision Tree (DT)

DT was implemented using *DecisionTreeRegressor* as a baseline nonlinear model. It recursively partitions the feature space to minimize prediction error, but is more prone to overfitting than ensemble methods. To somehow enhance the generalization ability of the model and prevent overfitting, a grid search was performed to optimize the key hyperparameters, including the maximum depth (*max_depth* $\in \{3, 5, 7, 10\}$), the minimum number of samples required for a leaf node (*min_samples_leaf* $\in \{1, 2, 4\}$), the splitting criterion (*criterion* $\in$ {'squared_error', 'friedman_mse'}), and the maximum number of features considered for each split (*max_features* $\in$ {'sqrt', 'log2', None}).

### e) Long Short-Term Memory (LSTM)

As an improved architecture of Recurrent Neural Networks (RNNs), LSTM effectively mitigates the gradient vanishing problem by introducing a gating mechanism, enabling it to capture long-term dependencies and dynamic patterns in meteorological time series. The model inputs are multivariate time series segments constructed via a sliding window, where each time step incorporates features including air temperature, humidity, precipitation, wind speed, sea level pressure, cloud cover, and solar radiation; the output layer consists of two neurons, corresponding to the predicted values of air temperature and humidity at future time steps, respectively.

In terms of model optimization, the LSTM architectures comprising different numbers of hidden layers {1, 2} with 50 or 100 units, coupled with the ReLU activation function, were evaluated. The Adam optimizer was employed for the training process, with the initial learning rate set to 0.001. Additionally, an early stopping mechanism was enabled: training is terminated when the validation loss fails to improve for 10 consecutive iterations, thereby preventing overfitting.

### f) CNN-LSTM

A hybrid deep learning architecture, Convolutional Neural Network–Long Short-Term Memory (CNN-LSTM), was constructed for multi-step prediction of air temperature and relative humidity. This model combines the local feature extraction capability of CNN with the time series modeling advantages of LSTM: first, a one-dimensional convolutional layer (1D-CNN) acts on the multivariate time series inputs constructed by a sliding window, automatically extracting local patterns and key temporal features of each variable in the time dimension; subsequently, the LSTM layer performs dynamic modeling on the high-dimensional feature sequences output by the convolution, effectively capturing long-term dependencies. Finally, the fully connected layer maps the hidden state of the LSTM to a dual-target output (air temperature and humidity).

After model optimization with the grid search strategy, the chosen CNN-LSTM architecture consists of a one-dimensional convolutional layer with 32 filters and a kernel size of 3, followed by an LSTM hidden layer with 50 units and a ReLU activation function. The Adam optimizer (initial learning rate of 0.001) was adopted for the training process, and an early stopping mechanism was enabled: training is terminated when the validation loss does not improve for 10 consecutive iterations.

### g) eXtreme Gradient Boosting (XGBoost)

As a highly efficient ensemble learning method based on Gradient Boosting Decision Trees (GBDT), XGBoost constructs multiple weak learners (regression trees) in a serial manner and continuously corrects residuals. Endowed with the advantages of high prediction accuracy, strong robustness, and a built-in regularization mechanism, it is particularly suitable for processing high-dimensional and nonlinear meteorological data. To realize the joint prediction for dual targets, the XGBoost regressor was encapsulated with MultiOutputRegressor in this study, enabling the model to output the predicted values of air temperature and relative humidity synchronously.

In terms of model optimization, grid search combined with time series cross-validation was adopted. The search space covered the number of trees (*n_estimators* $\in \{10, 50, 100\}$), the maximum tree depth (*max_depth* $\in \{3, 5, 7\}$), the learning rate (*learning_rate* $\in \{0.01, 0.1, 0.2\}$), the subsample ratio (*subsample* $\in \{0.7, 0.9\}$), the column sample ratio per tree

(*colsample_bytree* ∈ {0.7, 0.9}), and the minimum loss reduction required for a split (*gamma* ∈ {0, 0.1, 0.2}).

*D. Model Evaluation*

All models were evaluated for performance on both the training set and the test set. Four quantitative metrics were employed, with the arithmetic mean calculated for the scores of the two output variables (air temperature and relative humidity): mean absolute error (MAE); root mean square error (RMSE); coefficient of determination ($R^2$); and mean absolute percentage error (MAPE). To ensure numerical stability, an infinitesimal constant ($\epsilon$=np.finfo(np.float64).eps) was introduced in the MAPE calculation to avoid division-by-zero errors. In addition, visual diagnostic methods were adopted to support the evaluation, including time series comparisons and scatter plots of predicted and actual values for qualitative analysis of prediction performance.

*E. Implementation Details*

The experimental environment was built on Python 3.9. Conventional machine learning models, including RF, SVR, Multilayer Perceptron, and DT, were implemented using scikit-learn, while XGBoost was implemented using the XGBoost library with a scikit-learn-compatible interface. Their hyperparameters were optimized using *GridSearchCV* combined with TimeSeriesSplit (*k*=5), with *n_jobs*=-1 enabled for parallel computation. The deep learning models, namely LSTM and CNN-LSTM, were implemented in TensorFlow/Keras. Models were trained using a chronological 80/20 split, where the first 80% of the samples were used for training and the remaining 20% for testing. All figures and charts were generated using Matplotlib and Seaborn toolkits.

## IV. RESULTS

*A. Overall Model Performance*

This study systematically evaluates the performance of seven machine learning and deep learning models for air temperature and relative humidity prediction. As shown in **Table I**, for the air temperature prediction task, the XGBoost model achieves the best performance, with a test MAE of 0.302 °C, RMSE of 0.393 °C, $R^2$ of 0.995, and MAPE of only 2.884%, significantly outperforming other models. The SVR model ranks second, with a test $R^2$ of 0.994, showing performance close to that of XGBoost. MLP and RF also perform well, although RF shows a greater drop in test performance, suggesting some overfitting. In contrast, DT and LSTM produce noticeably larger test errors. CNN-LSTM performs worst on this task, with a test MAE of 3.230 °C and an $R^2$ of only 0.526, indicating that the current hybrid architecture does not adapt well to the structure and scale of the spatiotemporal features in the present dataset.

For the relative humidity prediction task (see **Table II**), XGBoost again demonstrates the strongest generalization ability, with a test MAE of 1.271%, RMSE of 2.116%, $R^2$ of 0.984, and MAPE of only 1.675%. The SVR model also performs excellently, with a test $R^2$ of 0.979, slightly inferior to XGBoost, followed by MLP and RF. Although RF maintains relatively high predictive accuracy, its larger RMSE suggests lower stability than XGBoost. DT remains the weakest conventional model for humidity forecasting, with a test MAE of 4.473% and an $R^2$ of 0.875. LSTM outperforms DT but still lags behind the best tree-based and kernel-based methods. CNN-LSTM yields moderate humidity performance, but it remains clearly inferior to XGBoost and SVR, further indicating that its structure may not fully adapt to the dynamic characteristics of humidity variations.

**Table III** summarizes the average performance of each model across the temperature and humidity tasks. XGBoost ranks first for all averaged metrics, with a test MAE of 0.787, RMSE of 1.254, $R^2$ of 0.989, and MAPE of 2.280%, indicating the strongest overall predictive capability. SVR follows closely, while MLP also remains competitive. In contrast, CNN-LSTM shows the weakest overall performance, with an average test MAE of 3.278, $R^2$ of 0.731, and MAPE of 18.215%. Overall, the results suggest that XGBoost provides the most effective balance of predictive accuracy and robustness for the present multivariate meteorological forecasting task.

TABLE I. PERFORMANCE COMPARISON OF MODELS ON THE TEMPERATURE PREDICTION TASK

| Model | Temperature Prediction Metrics | | | |
|---|---|---|---|---|
| | Test MAE | Test RMSE | Test $R^2$ | Test MAPE |
| SVR | 0.307 | 0.415 | 0.994 | 3.002% |
| MLP | 0.390 | 0.512 | 0.991 | 3.615% |
| RF | 0.630 | 0.877 | 0.974 | 5.852% |
| DT | 1.700 | 2.222 | 0.834 | 15.396% |
| LSTM | 1.520 | 1.909 | 0.877 | 17.060% |
| CNN-LSTM | 3.230 | 3.752 | 0.526 | 32.403% |
| XGBoost | 0.302 | 0.393 | 0.995 | 2.884% |

TABLE II. PERFORMANCE COMPARISON OF MODELS ON THE HUMIDITY PREDICTION TASK

| Model | Humidity Prediction Metrics | | | |
|---|---|---|---|---|
| | Test MAE | Test RMSE | Test $R^2$ | Test MAPE |
| SVR | 1.564 | 2.378 | 0.979 | 1.971% |
| MLP | 1.533 | 2.303 | 0.981 | 1.930% |
| RF | 1.733 | 2.721 | 0.973 | 2.237% |
| DT | 4.473 | 5.826 | 0.875 | 5.480% |
| LSTM | 2.580 | 3.063 | 0.966 | 3.000% |
| CNN-LSTM | 3.327 | 4.180 | 0.936 | 4.027% |
| XGBoost | 1.271 | 2.116 | 0.984 | 1.675% |

TABLE III. AVERAGE PERFORMANCE METRICS OF TEMPERATURE AND HUMIDITY PREDICTION TASKS

| Model | Average Metrics | | | |
|---|---|---|---|---|
| | Test MAE | Test RMSE | Test $R^2$ | Test MAPE |
| SVR | 0.936 | 1.397 | 0.987 | 2.487% |
| MLP | 0.962 | 1.407 | 0.986 | 2.773% |
| RF | 1.182 | 1.799 | 0.973 | 4.044% |
| DT | 3.086 | 4.024 | 0.854 | 10.438% |
| LSTM | 2.050 | 2.486 | 0.921 | 10.030% |
| CNN-LSTM | 3.278 | 3.966 | 0.731 | 18.215% |
| XGBoost | 0.787 | 1.254 | 0.989 | 2.280% |

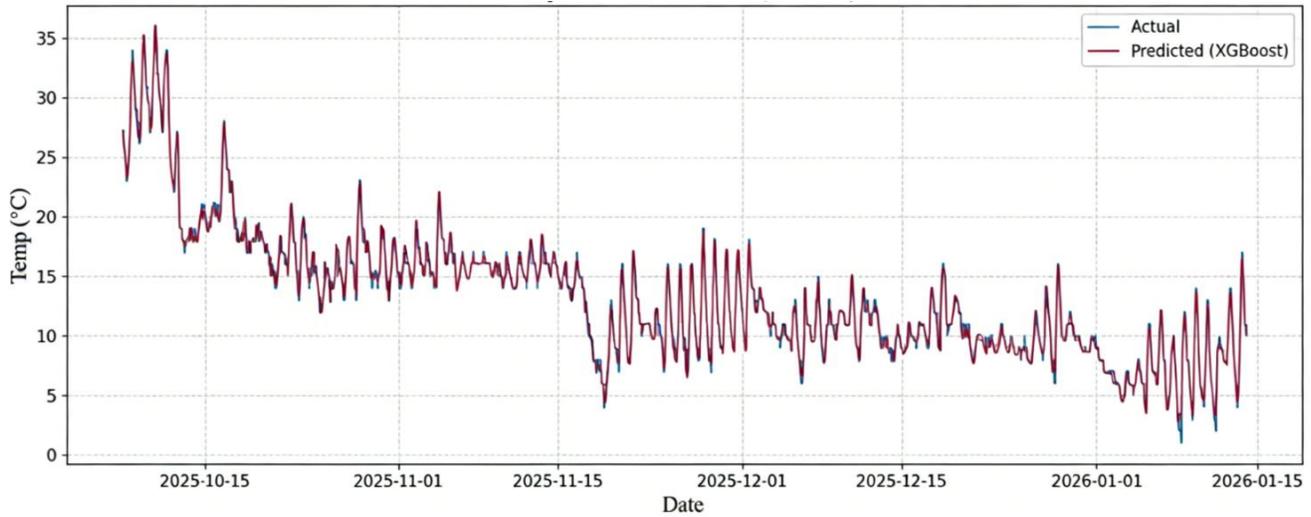

Figure 4.  Time series comparison of actual and predicted temperature on the test set by XGBoost.

## B. Visual Analytics

The visualization results further illustrate the predictive performance of the XGBoost model on hourly meteorological data in Chongqing. **Fig. 4** illustrates the time series comparison for the XGBoost model on the test set, where the predicted temperature curve (purple) closely tracks the actual observations (blue). The model captures both the broader cooling trend from October to January and the short-term variations in the data. The strong agreement between the predicted and observed curves suggests that XGBoost can effectively learn the temporal variation patterns of temperature.

The scatter plots in **Fig. 5** and **Fig. 6** further support the strong predictive performance of XGBoost for both temperature and humidity. In **Fig. 5**, the predicted temperature values are tightly distributed around the ideal 1:1 line, with only limited dispersion, indicating high pointwise agreement between predictions and observations. Similarly, **Fig. 6** shows that the predicted humidity values also cluster closely around the diagonal reference line, demonstrating strong predictive consistency across the observed humidity range. These visual results are consistent with the quantitative metrics reported in **Tables I** and **II**, where XGBoost achieves the best overall test performance, including the lowest MAE and RMSE, and the highest R² among all compared models. For comparison, **Fig. 9** and **Fig. 10** present the corresponding scatter plots for CNN-LSTM. Relative to XGBoost, the CNN-LSTM predictions exhibit noticeably larger dispersion from the 1:1 line, especially for temperature, indicating weaker pointwise agreement and lower predictive accuracy. This observation is consistent with the quantitative results in **Tables I–III**, where CNN-LSTM shows substantially poorer test performance than XGBoost and the other competitive baselines.

Overall, the visual and quantitative results consistently indicate that XGBoost provides the strongest overall performance for the dual-target forecasting task considered in this study. Whereas CNN-LSTM shows clearly weaker generalization performance under the current experimental setting.

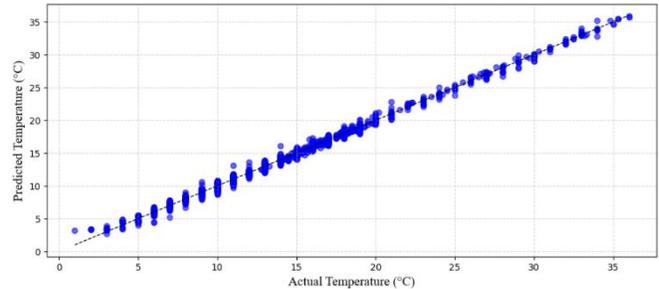

Figure 5.  Scatter plot of actual versus predicted temperature on the test set with the ideal 1:1 line (XGBoost).

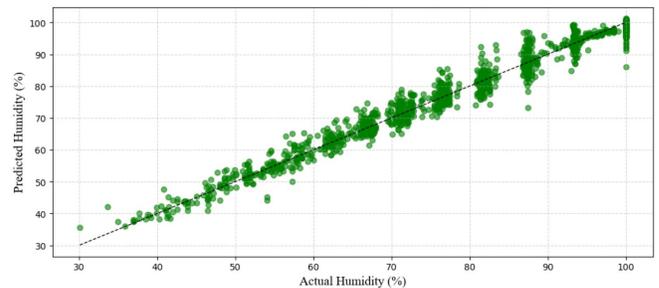

Figure 6.  Scatter plot of actual versus predicted humidity on the test set with the ideal 1:1 line (XGBoost).

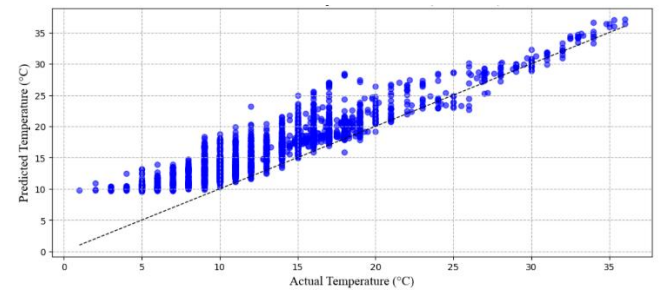

Figure 7.  Scatter plot of actual versus predicted temperature on the test set with the ideal 1:1 line (CNN-LSTM)

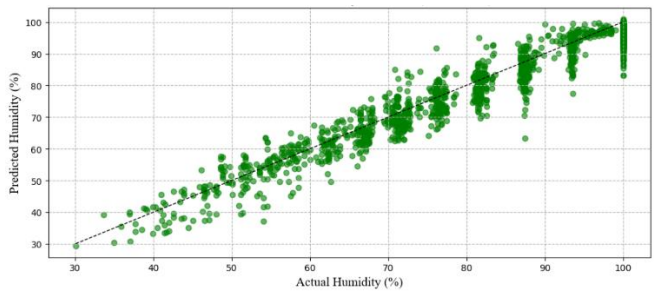

Figure 8. Scatter plot of actual versus predicted humidity on the test set with the ideal 1:1 line (CNN-LSTM)

## V. Discussion

The results of this study indicate that XGBoost achieves the best performance across all quantitative metrics for both temperature and humidity forecasting using hourly meteorological data from Chongqing. Its strong generalization ability is likely related to its effectiveness in handling nonlinear relationships, heterogeneous features, and relatively high-dimensional structured inputs, while maintaining resistance to overfitting. This finding is broadly consistent with previous studies showing the competitiveness of boosting-based models in data-driven prediction tasks in other domains, e.g., [6], [7]. By contrast, inconsistent with [8], [9], the deep learning models considered here, LSTM and CNN-LSTM, do not outperform ensemble methods in this study under the current data scale and modeling setup. CNN-LSTM performs the worst, as its temporal feature extraction ability is limited by the current network structure and data scale, which aligns with the finding that such architectures need larger datasets or more sophisticated tuning to surpass simpler models [10].

The visual analysis further supports the quantitative results. The XGBoost time-series and scatter plots show strong agreement between predictions and observations for both temperature and humidity. In contrast, the CNN-LSTM scatter plots exhibit noticeably larger dispersion from the ideal 1:1 line, particularly for temperature prediction, which is consistent with its substantially lower $R^2$ and larger test errors. These findings suggest that the additional model complexity of CNN-LSTM does not necessarily translate into improved forecasting performance in the present task.

From a practical perspective, XGBoost can be recommended as an effective model for short-term meteorological forecasting when predictive accuracy and robustness are the primary objectives. SVR and MLP also provide competitive alternatives, while RF remains a useful baseline. Future work may further investigate improved deep learning architectures, incorporate spatial information from neighboring stations [9], [11], and explore physics-informed machine learning [12] and semi-supervised learning techniques [13] to enhance generalization and interpretability in meteorological forecasting.

## VI. Conclusion

This study compared seven machine learning and deep learning models for hourly forecasting of air temperature and relative humidity in Chongqing, a mountainous city with complex microclimatic conditions. The results show that XGBoost achieves the best overall predictive performance across all quantitative metrics, with a test MAE of 0.302 °C for temperature and 1.271% for humidity, and an average $R^2$ of 0.989. The visual results are consistent with the quantitative evaluation, showing strong agreement between XGBoost predictions and the actual observations, while CNN-LSTM exhibits substantially larger deviations and weaker accuracy under the current experimental setting. Overall, XGBoost provides the most effective solution for the dual-target forecasting task. Future work will focus on improved deep learning architectures, the incorporation of spatial information from neighboring stations [9], [11], and physics-informed forecasting approaches [12]. In addition, semi-supervised learning may be explored to address the limited availability of labeled meteorological data and to better capture spatiotemporal correlations [13].

## References

[1] S. Feng, X. Gong, and Y. Zeng, "Machine learning methods for sub-seasonal to seasonal prediction of temperature and precipitation over China," Advances in Atmospheric Sciences, vol. 38, no. 9, pp. 1455–1470, Sep. 2021.

[2] H. V. Moosavi, M. R. Delavar, and A. Mohammadzadeh, "A comparative study of machine learning models for short-term air temperature forecasting," Atmospheric Research, vol. 228, pp. 243–256, Oct. 2019.

[3] T. Chen and C. Guestrin, "XGBoost: A scalable tree boosting system," in Proc. 22nd ACM SIGKDD Int. Conf. Knowl. Discov. Data Min., San Francisco, CA, USA, Aug. 2016, pp. 785–794.

[4] S. Hochreiter and J. Schmidhuber, "Long short-term memory," Neural Computation, vol. 9, no. 8, pp. 1735–1780, Nov. 1997.

[5] M. A. De Pinto, M. L. Delle Monache, and S. Speranza, "Hybrid CNN-LSTM model for short-term weather forecasting," in Proc. Int. Conf. Comput. Sci. Comput. Intell., Dubai, UAE, Dec. 2020, pp. 112–117.

[6] Y. Dong, K. Chen, and Z. Ma, "Comparative Study on Semi-Supervised Learning Applied for Anomaly Detection in Hydraulic Condition Monitoring System," in Proc. IEEE Int. Conf. Syst., Man, Cybern. (SMC), Honolulu, Oahu, HI, USA, 2023, pp. 1702–1708, doi: 10.1109/SMC53992.2023.10394193.

[7] Y. Dong, K. Chen, Y. Peng, and Z. Ma, "Comparative Study on Supervised versus Semi-supervised Machine Learning for Anomaly Detection of In-vehicle CAN Network," in Proc. IEEE 25th Int. Conf. Intell. Transp. Syst. (ITSC), Macau, China, 2022, pp. 2914–2919, doi: 10.1109/ITSC55140.2022.9922235.

[8] B. Wang, J. Lu, Z. Wang, and G. Zhang, "A survey on deep learning for time series forecasting," Neurocomputing, vol. 451, pp. 116–137, Sep. 2021.

[9] Y. Huang, Y. Dong, Y. Tang, and L. Li, "Parking Availability Prediction via Fusing Multi-Source Data with A Self-Supervised Learning Enhanced Spatio-Temporal Inverted Transformer," arXiv preprint arXiv:2509.04362, Sep. 2025. [Online]. Available: https://doi.org/10.48550/arXiv.2509.04362

[10] Y. LeCun, Y. Bengio, and G. Hinton, "Deep learning," Nature, vol. 521, no. 7553, pp. 436–444, May 2015.

[11] Y. Huang, Y. Dong, Y. Tang, and L. Li, "Leverage multi-source traffic demand data fusion with transformer model for urban parking prediction," in Proc. 28th Int. Conf. Hong Kong Soc. Transp. Stud. (HKSTS), 2024, preprint available at https://arxiv.org/abs/2405.01055.

[12] A. Karpatne et al., "Theory-guided data science: A new paradigm for scientific discovery from data," IEEE Trans. Knowl. Data Eng., vol. 29, no. 10, pp. 2318–2331, Oct. 2017.

[13] Y. Huang, Y. Dong, Y. Tang, and A. García-Hernandez, "A self-supervised transformer for unusable shared bike detection," in 2025 IEEE 28th Int. Conf. Intell. Transp. Syst. (ITSC), Gold Coast, Australia, 2025, pp. 577–582, doi: 10.1109/ITSC60802.2025.11423728.